\pdfoutput=1

\documentclass[11pt]{article}

\usepackage[]{acl}

\usepackage{times}
\usepackage{latexsym}
\usepackage{booktabs}
\usepackage{multirow}
\usepackage[T1]{fontenc}

\usepackage[utf8]{inputenc}

\usepackage{microtype}

\usepackage{graphicx}

\title{Exploring the Feasibility of ChatGPT for Event Extraction}

\author{ Jun Gao\textsuperscript{1}\thanks{Work done when Jun Gao was interning at 4Paradigm.} \enskip Huan Zhao\textsuperscript{2} \enskip Changlong Yu\textsuperscript{3} \enskip Ruifeng Xu\textsuperscript{1}\\
 \textsuperscript{1}Harbin Institute of Technology (Shenzhen)\\
   \textsuperscript{2}4Paradigm. Inc.\quad \textsuperscript{3}HKUST, Hong Kong, China  \\
\normalsize \texttt{imgaojun@gmail.com} \enskip \texttt{zhaohuan@4paradigm.com} \enskip \texttt{cyuaq@cse.ust.hk}
}

\begin{document}
\maketitle

\begin{abstract}
  Event extraction is a fundamental task in natural language processing that involves identifying and extracting information about events mentioned in text. However, it is a challenging task due to the lack of annotated data, which is expensive and time-consuming to obtain. The emergence of large language models (LLMs) such as ChatGPT provides an opportunity to solve language tasks with simple prompts without the need for task-specific datasets and fine-tuning. While ChatGPT has demonstrated impressive results in tasks like machine translation, text summarization, and question answering, it presents challenges when used for complex tasks like event extraction. Unlike other tasks, event extraction requires the model to be provided with a complex set of instructions defining all event types and their schemas. To explore the feasibility of ChatGPT for event extraction and the challenges it poses, we conducted a series of experiments. Our results show that ChatGPT has, on average, only 51.04\% of the performance of a task-specific model such as EEQA in long-tail and complex scenarios. Our usability testing experiments indicate that ChatGPT is not robust enough, and continuous refinement of the prompt does not lead to stable performance improvements, which can result in a poor user experience. Besides, ChatGPT is highly sensitive to different prompt styles.
   \end{abstract}

\section{Introduction}
Event extraction is a crucial natural language processing task that involves identifying and extracting structured events mentioned in text~\citep{Du2020EventEB,Lu2021Text2EventCS}. However, the limited availability of annotated data makes this task particularly challenging, as expert annotation can be expensive~\citep{Yang2019ExploringPL,Lou2022TranslationBasedIA,gao-etal-2022-mask}. Recent advancements in large language models (LLMs), such as ChatGPT~\footnote{\url{https://chat.openai.com/chat}}, have generated considerable interest in applying these models to various language tasks, such as machine translation~\citep{Jiao2023IsCA}, text summarization~\citep{Bang2023AMM}, and automatic bug fixing~\citep{Sobania2023AnAO}, requiring only a few demonstrations.

\begin{figure}[t]
  \centering
  \includegraphics[width=0.9\linewidth]{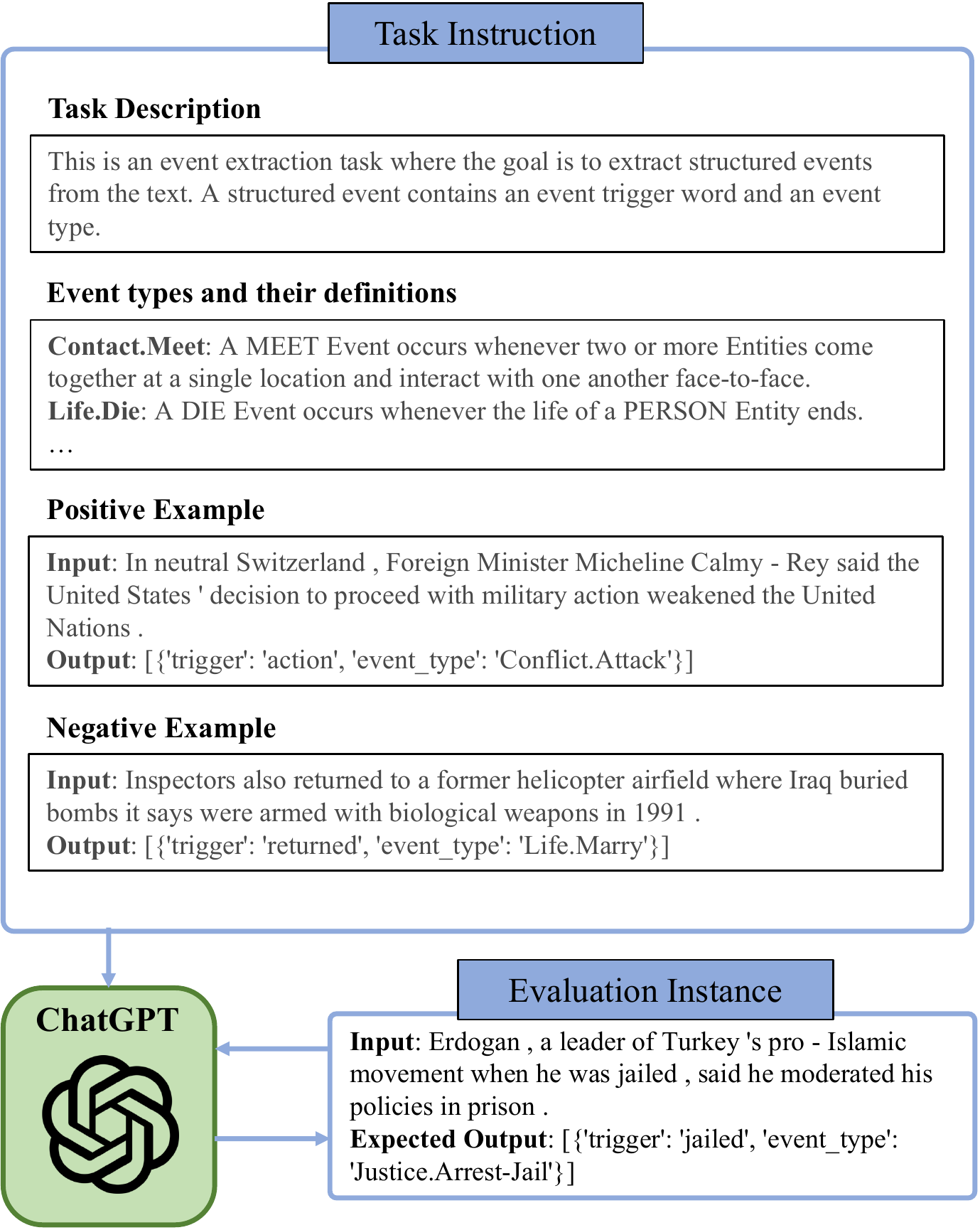}
  \caption{An example of zero-shot event extraction using ChatGPT. In this scenario, ChatGPT is tasked with processing a pool of evaluation instances based on provided instructions, which include the task description, event type definitions and demonstration examples. ChatGPT is expected to produce responses for each evaluation instance without any prior training on the specific task or data.}
  \label{fig:example}
\end{figure}

However, event extraction is a more complex task that poses unique challenges. Unlike other tasks like translation and summarization, event extraction requires a more detailed and nuanced description of the task. To utilize ChatGPT for event extraction, the model must be provided with a schema for each event type, including the triggers and arguments to be extracted. Additionally, handling complex special cases is critical for accurate event extraction.

This paper aims to explore the feasibility of ChatGPT for event extraction. We conduct experiments to evaluate ChatGPT's performance compared to two task-specific models such as Text2Event~\citep{Lu2021Text2EventCS} and EEQA~\citep{Du2020EventEB} in long-tail and complex scenarios~(i.e., texts containing multiple events). Our results show that ChatGPT has, on average, only 51.04\% of the performance of EEQA in such scenarios. We also conduct usability testing experiments to assess whether users can achieve their desired outcomes with minimal attempts. Our experiments indicate that ChatGPT is not robust enough. Despite several prompt optimizations aimed at enhancing its performance on the event extraction task, ChatGPT did not exhibit a consistent upward trend. This limitation can result in a poor user experience. Additionally, ChatGPT is highly sensitive to different prompt styles, resulting in significant variations in the results obtained by different users using ChatGPT.

\section{Background}

\paragraph{Event definition.}
A universally agreed definition for events is currently lacking, as definitions may vary depending on the application or task~\citep{Liu2020ExtractingEA}. This paper will use the definition provided by the Automatic Content Extraction (ACE) 2005 evaluation~\footnote{\url{https://catalog.ldc.upenn.edu/LDC2006T06}}, which defines an event as ``a specific occurrence involving participants,'' and includes several arguments with different roles. Within this framework, an \textbf{event mention} is a sentence that describes an event, an \textbf{event trigger} is the word that most clearly expresses the event, and \textbf{event arguments} are taggable entities involved in the event. Typically, specific participant roles can be filled for each type of event. For instance, in the sentence "A number of demonstrators threw stones and empty bottles at Israeli soldiers positioned near a Jewish holy site at the town's entrance," the event described is an ATTACK, with the entire sentence serving as an event mention, "threw" acting as the event trigger, and four event arguments, including the ATTACK-Attacker (demonstrators), ATTACK-Target (Israeli soldiers), ATTACK-Instrument (stones and empty bottles), and ATTACK-Place (a Jewish holy site at the town's entrance).

\paragraph{Event Extraction Tasks.}
Event extraction refers to the process of obtaining structured information from unstructured text. In the context of ACE 2005, event extraction consists of two subtasks. The first subtask, known as \textbf{event detection}, involves identifying events and classifying them according to their specific types. The second subtask, \textbf{event argument extraction}, aims to identify and extract the various words, phrases, and entities that fulfill different roles within a given event. In this work, we mainly focus on the event detection task.

\section{ChatGPT for Event Extraction}
Event extraction aims to identify event triggers and their corresponding arguments from unstructured text and present them in a structured format for further processing. In this context, our objective is to evaluate ChatGPT's ability to perform zero-shot event extraction without fine-tuning the model.
To leverage the advanced capabilities of the ChatGPT model for event extraction, a promising approach is to frame the task as a multi-turn question answering problem. This method builds upon prior research~\citep{Li2019EntityRelationEA,Du2020EventEB} and involves using the model to identify events by posing a series of questions in a conversational style. 

Specifically, given a task instruction that describes this task in natural language, ChatGPT is expected to output the events in the text and represent them in a structured form. In this work, we mainly focus on the event detection task, so that each event output contains an event trigger word and an event type, and we use the JSON format to serialize the output because it is widely used and easy to parse.
Figure~\ref{fig:example} provides An example of zero-shot event extraction using ChatGPT. In this scenario, ChatGPT is tasked with processing a pool of evaluation instances based on provided instructions, which include the task description, event type definitions and demonstration examples. ChatGPT is expected to produce responses for each evaluation instance without any prior training on the specific task or data.

In this study, we first assess the performance of ChatGPT in comparison to task-specific models~(Text2Event and EEQA) under real-world conditions. Specifically, we randomly select 20 samples from the raw test set to evaluate the efficacy of ChatGPT. Then, we conduct an analysis of the impact of different elements in the prompt on the performance of ChatGPT using the same 20 test samples. To obtain a comprehensive understanding of ChatGPT's performance, we also evaluate its performance in both long-tail and complex scenarios (i.e., texts containing multiple events) in comparison to task-specific models. Finally, we conduct usability testing to assess ChatGPT's user-friendliness for the event extraction task.

\section{Experimental Setup and Results}

\subsection{Setup}

\paragraph{Dataset.}
We conduct experiments on the ACE 2005 corpus, which comprises documents crawled between 2003 and 2005 from diverse sources, including newswire, weblogs, broadcast conversations, and broadcast news. To maintain consistency with prior works~\citep{Du2020EventEB,Lu2021Text2EventCS,gao-etal-2022-mask}, we employ the same data split and preprocessing steps.

\paragraph{Evaluation Metric.}
We utilize the same criteria defined in \citet{Du2020EventEB} for our analysis. Specifically, an event trigger is considered correctly identified (ID) if its offsets correspond to those of a gold-standard trigger. Moreover, for a trigger to be correctly classified, its event type (there are 33 types in total) must match that of the gold-standard trigger.

\paragraph{Baselines.} 
In our experiments, we conduct a comparison between ChatGPT and several event extraction baselines, including Text2Event and EEQA.
\begin{itemize}
  \item \textbf{Text2Event}~\citep{Lu2021Text2EventCS}: Text2Event is a framework that utilizes T5 models~\citep{Raffel2019ExploringTL} to approach event extraction by framing it as a SEQ2SEQ generation task. In this method, all triggers, arguments, and their corresponding labels are generated as natural language words.
  \item \textbf{EEQA}~\citep{Du2020EventEB}: EEQA, on the other hand, approaches event extraction by formulating it as a question answering task. They developed two BERT-based QA models - one for detecting event triggers and another for extracting arguments. In our experiments, we solely utilized the BERT-based QA model for event trigger detection.
\end{itemize}

\subsection{Results}
\paragraph{Comparison with task-specific models.}
We first compare the performance of ChatGPT with two task-specific models, namely EEQA and Text2Event, which are fine-tuned using the ACE05 training set. To assess the models, we randomly select 20 samples from the ACE05 test set. Table~\ref{tab:main} displays the results obtained from the experiment. The findings reveal that EEQA achieves the highest F1 performance, while ChatGPT lags behind Text2Event and EEQA. Notably, ChatGPT's Recall is comparable to Text2Event (T5-base), but its precision is significantly lower. Upon analyzing the output samples, we observe that ChatGPT extracted more event triggers, potentially because it lacks a clear understanding of certain event definitions.

\begin{table}[tbp]
  \begin{tabular}{@{}lccc@{}}
  \toprule
  \multirow{2}{*}{} & \multicolumn{3}{c}{\textbf{Event Detection}} \\ \cmidrule(l){2-4} 
   & \textbf{P} & \textbf{R} & \textbf{F1} \\ \midrule
  ChatGPT & 57.14 & 72.73 & 64.00 \\
  Text2Event (T5-base) & 75.76 & 75.76 & 75.76 \\
  Text2Event (T5-large) & \textbf{82.76} & 72.73 & 77.42 \\
  EEQA (BERT-base) & 81.82 & \textbf{81.82} & \textbf{81.82} \\ \bottomrule
  \end{tabular}
  \caption{Comparison with task-specific models.}
  \label{tab:main}
  \end{table}
  
  \begin{table}[tbp]
    \begin{tabular}{@{}lccc@{}}
    \toprule
    \multirow{2}{*}{} & \multicolumn{3}{c}{\textbf{Event Detection}} \\ \cmidrule(l){2-4} 
     & \textbf{P} & \textbf{R} & \textbf{F1} \\ \midrule
    Full Prompt & 57.14 & 72.73 & 64.00 \\
    - Positive Example & 48.00 & 72.73 & 57.83 \\
    - Negative Example & 62.50 & 75.76 & 68.49 \\
    - Event Type Definition & 47.50 & 57.58 & 52.05 \\ \bottomrule
    \end{tabular}
    \caption{Instructing with different elements.}
    \label{tab:ablation}
    \end{table}

\paragraph{Instructing with different elements.}

\begin{table*}[htbp]
  \centering
  \small
  \begin{tabular}{@{}lllllllllllll@{}}
  \toprule
  \multirow{2}{*}{} & \multicolumn{3}{c}{\textbf{High Frequency}} & \multicolumn{3}{c}{\textbf{Low Frequency}} & \multicolumn{3}{c}{\textbf{Simple Examples}} & \multicolumn{3}{c}{\textbf{Complex Examples}} \\ \cmidrule(l){2-13} 
   & \multicolumn{1}{c}{\textbf{P}} & \multicolumn{1}{c}{\textbf{R}} & \multicolumn{1}{c}{\textbf{F1}} & \multicolumn{1}{c}{\textbf{P}} & \multicolumn{1}{c}{\textbf{R}} & \multicolumn{1}{c}{\textbf{F1}} & \multicolumn{1}{c}{\textbf{P}} & \multicolumn{1}{c}{\textbf{R}} & \multicolumn{1}{c}{\textbf{F1}} & \multicolumn{1}{c}{\textbf{P}} & \multicolumn{1}{c}{\textbf{R}} & \multicolumn{1}{c}{\textbf{F1}} \\ \midrule
  ChatGPT & 23.08 & 45.00 & 30.51 & 23.40 & 55.00 & 32.84 & 49.00 & 50.00 & 28.99 & 42.31 & 52.38 & 46.81 \\
  Text2Event (t5-base) & 55.56 & 50.00 & 52.63 & \textbf{61.90} & \textbf{65.00} & \textbf{63.41} & \textbf{79.17} & \textbf{95.00} & \textbf{86.36} & 81.82 & 64.29 & 72.00 \\
  Text2Event (t5-large) & 54.55 & 60.00 & 57.14 & 60.00 & 60.00 & 60.00 & 75.00 & 90.00 & 81.82 & \textbf{88.57} & \textbf{73.81} & \textbf{80.52} \\
  EEQA & \textbf{59.09} & \textbf{65.00} & \textbf{61.90} & 61.11 & 55.00 & 57.89 & 78.26 & 90.00 & 83.72 & 82.35 & 66.67 & 73.68 \\ \bottomrule
  \end{tabular}
  \caption{Performance of ChatGPT and task-specific models in long-tail and complex scenarios.}
  \label{tab:analysis}
  \end{table*}

The prompt plays a critical role in enabling ChatGPT to perform event extraction, as it contains vital information, such as task descriptions, event type definitions, and demonstrations. However, the impact of each of these elements on ChatGPT's performance in event extraction remains unclear. To address this gap, we conduct an ablation experiment to examine the effect of different prompt elements on ChatGPT's performance.
To this end, we design four prompts, each with varying amounts and types of information. The prompts are as follows:
\begin{itemize}
  \item Full Prompt: The complete prompt, including task description, event type definitions, and demonstrations.
  \item No Event Definitions: The prompt without the event definitions.
  \item No Positive Example: The prompt without the positive sample.
  \item No Negative Example: The prompt without the negative sample.
\end{itemize}
We then evaluate ChatGPT's performance on the event extraction task for each prompt, and we use the same 20 test samples used in the previous experiment
As shown in Table~\ref{tab:ablation}, there is a marked decline in model performance when the positive example and event type definitions are removed. Specifically, removing the positive example results in a 6.17-point decrease in F1, while removing the event type definitions leads to an 11.95-point decrease in F1. Interestingly, the model's performance improved after eliminating the negative sample. We speculate that this might be because the model can not comprehend the meaning of  ``negative sample'' and misinterpreted them as positive samples.
A similar finding can be found in the prior work~\citep{Wang2022SuperNaturalInstructionsGV}.

\paragraph{Can ChatGPT consistently output structured events?}
Since the goal of event extraction is to automatically extract structured events from text and then use them for downstream tasks, the event extraction model needs to be able to output structured events in a stable manner.
After analyzing the output of ChatGPT in the ablation experiment, we discover that the ChatGPT is able to consistently produce structured events in the vast majority of cases. Specifically, ChatGPT is able to produce structured events~(see Figure~\ref{fig:example} for example) for all 20 samples when provided with either the full prompt or the prompt without the positive example. However, when the negative sample and event type definitions are removed, only 19 out of 20 examples have a structured representation. This is because ChatGPT would output an explanation to indicate that it can not recognize the event when it encountered an unrecognizable event trigger word or event type, such as ``[{}] (There is no event trigger word or event type present in the input sentence.)''.

\paragraph{Analysis on long-tail and complex scenarios.}
To gain a more comprehensive understanding of ChatGPT's strengths and weaknesses, we conducted evaluations in two different scenarios, namely the long-tail scenario and the complex scenario.

\textbf{Long-tail scenario.} To evaluate ChatGPT's performance in handling low-frequency events, we split the test set into two categories based on event frequency. High-frequency events are those that appeared in at least 10 instances in the dataset, while low-frequency events are those that appeared less than 10 times. We randomly sample 20 test instances from each category and evaluate ChatGPT's performance separately on these two subsets of data. Our results in Table~\ref{tab:analysis} show that ChatGPT's performance on high-frequency and low-frequency events is not significantly different. However, ChatGPT's performance is significantly worse than the baseline models, achieving only 49.2\% and 56.7\% of EEQA's performance on low-frequency and high-frequency events, respectively.

\textbf{Complex scenario.} To assess ChatGPT's ability to handle complex samples, we divide the test set into two categories based on the number of events present in the text. Simple samples contain only one event, while complex samples contain more than two events. We randomly sample 20 test instances from each category and evaluate ChatGPT's performance separately on these two subsets of data. As shown in Table~\ref{tab:analysis}, ChatGPT's performance on both simple and complex samples is significantly worse than that of EEQA and Text2Event.

\begin{figure}[htbp]
  \centering
  \includegraphics[width=0.9\linewidth]{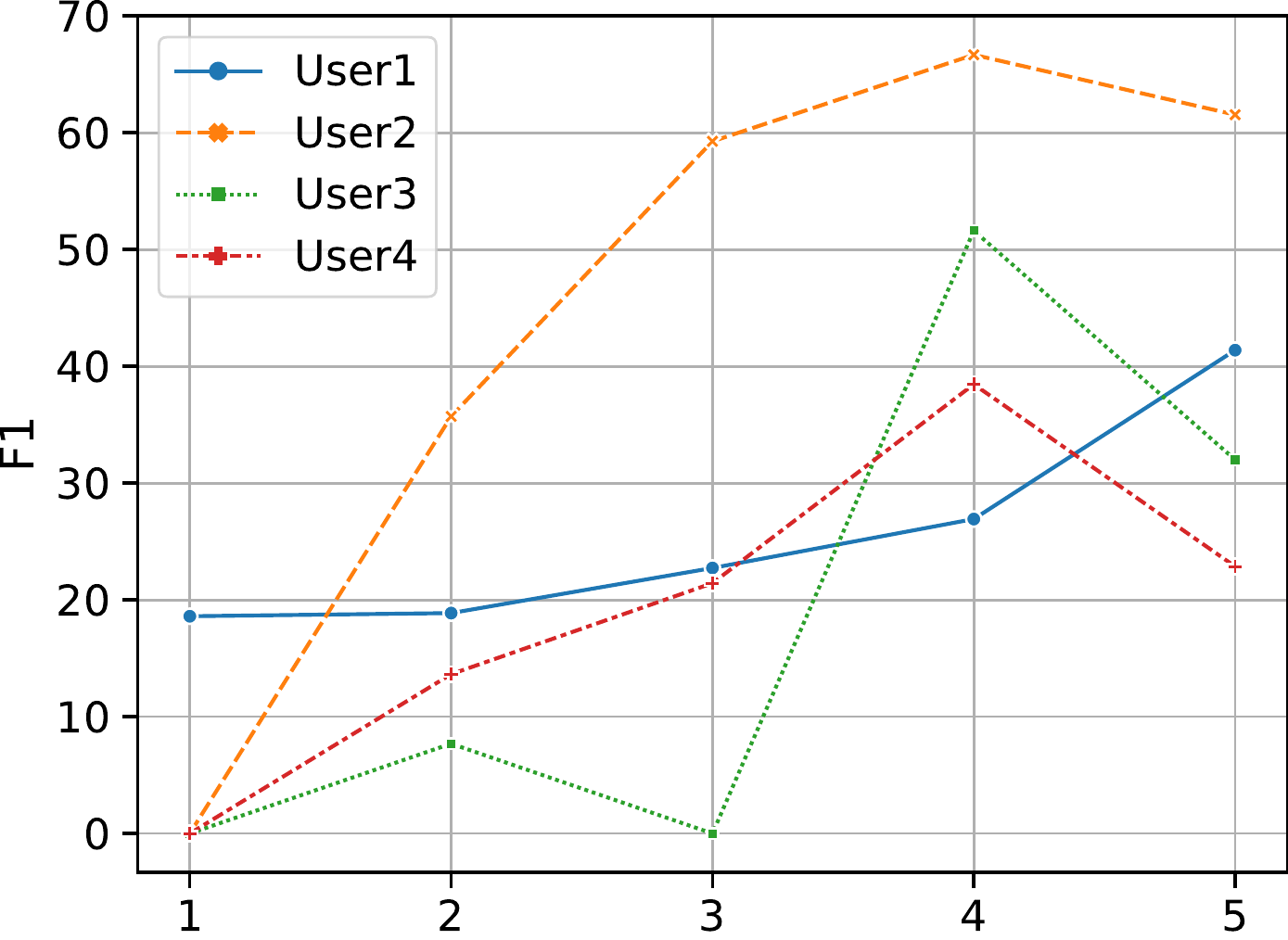}
  \caption{Performance of ChatGPT with different number of attempts by four annotators.}
  \label{fig:human}
\end{figure}

\paragraph{Usability of ChatGPT for event extraction.}
The ChatGPT requires a prompt to execute specific tasks. However, users currently lack guidance on creating suitable prompts. This section aims to evaluate ChatGPT's usability by assessing whether users can achieve their desired outcomes with minimal attempts.
To conduct the experiment, we recruited four professional and well-educated annotators~(e.g. postgraduate student on NLP research) to evaluate ChatGPT's usability. We randomly selected ten samples from the ACE05 test set and provided each annotator with five attempts to create a task prompt that would enable ChatGPT to extract structured events from the given text. Figure~\ref{fig:human} presents the results of the four annotators.
\begin{itemize}
  \item ChatGPT is not robust enough. Despite several prompt optimizations aimed at enhancing its performance on the event extraction task, ChatGPT did not exhibit a consistent upward trend. This limitation can result in a poor user experience.
  \item ChatGPT's performance is sensitive to different prompt styles. Different annotators utilized different prompt styles, resulting in significant performance variations for ChatGPT. For instance, ChatGPT achieved its highest F1 score of 66.67 on the fourth attempt, while the lowest is only 26.92.
\end{itemize}





\section{Conclusion}
In conclusion, our study highlights the potential of large language models like ChatGPT for event extraction, but also underscores the challenges associated with this task. Our findings suggest that, while ChatGPT can achieve good results in simple scenarios, it struggles to match the performance of task-specific models in more complex and long-tail scenarios. Our usability testing experiments also reveal that the performance of ChatGPT is highly dependent on the prompt style, which can lead to significant variations in results obtained by different users. These findings suggest that, while ChatGPT has enormous potential for event extraction, further research is needed to refine its capabilities and overcome the challenges it presents in this domain. Overall, our study contributes to the growing body of research on large language models and highlights the need for continued investigation into their strengths and limitations in complex NLP tasks like event extraction.

\section*{Acknowledgements}
We would like to express our sincere gratitude to Fangqi Zhu and Tianyi Xiao for their valuable participation in our human evaluation experiments. Their insightful comments and suggestions have greatly contributed to the quality of this paper.
We also extend our thanks to Wei Wang for his valuable feedback and discussion of the paper. His contributions have been instrumental in shaping our research.

\bibliography{anthology,custom}

\begin{thebibliography}{12}
\expandafter\ifx\csname natexlab\endcsname\relax\def\natexlab#1{#1}\fi

\bibitem[{Bang et~al.(2023)Bang, Cahyawijaya, Lee, Dai, Su, Wilie, Lovenia, Ji,
  Yu, Chung, Do, Xu, and Fung}]{Bang2023AMM}
Yejin Bang, Samuel Cahyawijaya, Nayeon Lee, Wenliang Dai, Dan Su, Bryan Wilie,
  Holy Lovenia, Ziwei Ji, Tiezheng Yu, Willy Chung, Quyet~V. Do, Yan Xu, and
  Pascale Fung. 2023.
\newblock A multitask, multilingual, multimodal evaluation of chatgpt on
  reasoning, hallucination, and interactivity.
\newblock \emph{ArXiv}, abs/2302.04023.

\bibitem[{Du and Cardie(2020)}]{Du2020EventEB}
X.~Du and Claire Cardie. 2020.
\newblock Event extraction by answering (almost) natural questions.
\newblock In \emph{Conference on Empirical Methods in Natural Language
  Processing}.

\bibitem[{Gao et~al.(2022)Gao, Yu, Wang, Zhao, and Xu}]{gao-etal-2022-mask}
Jun Gao, Changlong Yu, Wei Wang, Huan Zhao, and Ruifeng Xu. 2022.
\newblock \href {https://aclanthology.org/2022.findings-emnlp.332}
  {Mask-then-fill: A flexible and effective data augmentation framework for
  event extraction}.
\newblock In \emph{Findings of the Association for Computational Linguistics:
  EMNLP 2022}, pages 4537--4544, Abu Dhabi, United Arab Emirates. Association
  for Computational Linguistics.

\bibitem[{Jiao et~al.(2023)Jiao, Wang, tse Huang, Wang, and Tu}]{Jiao2023IsCA}
Wenxiang Jiao, Wenxuan Wang, Jen tse Huang, Xing Wang, and Zhaopeng Tu. 2023.
\newblock Is chatgpt a good translator? a preliminary study.
\newblock \emph{ArXiv}, abs/2301.08745.

\bibitem[{Li et~al.(2019)Li, Yin, Sun, Li, Yuan, Chai, Zhou, and
  Li}]{Li2019EntityRelationEA}
Xiaoya Li, Fan Yin, Zijun Sun, Xiayu Li, Arianna Yuan, Duo Chai, Mingxin Zhou,
  and Jiwei Li. 2019.
\newblock Entity-relation extraction as multi-turn question answering.
\newblock In \emph{Annual Meeting of the Association for Computational
  Linguistics}.

\bibitem[{Liu et~al.(2020)Liu, Chen, Liu, Zuo, and Zhao}]{Liu2020ExtractingEA}
Kang Liu, Yubo Chen, Jian Liu, Xinyu Zuo, and Jun Zhao. 2020.
\newblock Extracting events and their relations from texts: A survey on recent
  research progress and challenges.
\newblock \emph{AI Open}, 1:22--39.

\bibitem[{Lou et~al.(2022)Lou, Gao, Yu, Wang, Zhao, Tu, and
  Xu}]{Lou2022TranslationBasedIA}
Chenwei Lou, Jun Gao, Changlong Yu, Wei Wang, Huan Zhao, Weiwei Tu, and Ruifeng
  Xu. 2022.
\newblock Translation-based implicit annotation projection for zero-shot
  cross-lingual event argument extraction.
\newblock \emph{Proceedings of the 45th International ACM SIGIR Conference on
  Research and Development in Information Retrieval}.

\bibitem[{Lu et~al.(2021)Lu, Lin, Xu, Han, Tang, Li, Sun, Liao, and
  Chen}]{Lu2021Text2EventCS}
Yaojie Lu, Hongyu Lin, Jin Xu, Xianpei Han, Jialong Tang, Annan Li, Le~Sun,
  M.~Liao, and Shaoyi Chen. 2021.
\newblock Text2event: Controllable sequence-to-structure generation for
  end-to-end event extraction.
\newblock \emph{ArXiv}, abs/2106.09232.

\bibitem[{Raffel et~al.(2019)Raffel, Shazeer, Roberts, Lee, Narang, Matena,
  Zhou, Li, and Liu}]{Raffel2019ExploringTL}
Colin Raffel, Noam~M. Shazeer, Adam Roberts, Katherine Lee, Sharan Narang,
  Michael Matena, Yanqi Zhou, Wei Li, and Peter~J. Liu. 2019.
\newblock Exploring the limits of transfer learning with a unified text-to-text
  transformer.
\newblock \emph{ArXiv}, abs/1910.10683.

\bibitem[{Sobania et~al.(2023)Sobania, Briesch, Hanna, and
  Petke}]{Sobania2023AnAO}
Dominik Sobania, Martin Briesch, Carol Hanna, and Justyna Petke. 2023.
\newblock An analysis of the automatic bug fixing performance of chatgpt.
\newblock \emph{ArXiv}, abs/2301.08653.

\bibitem[{Wang et~al.(2022)Wang, Mishra, Alipoormolabashi, Kordi, Mirzaei,
  Arunkumar, Ashok, Dhanasekaran, Naik, Stap, Pathak, Karamanolakis, Lai,
  Purohit, Mondal, Anderson, Kuznia, Doshi, Patel, Pal, Moradshahi, Parmar,
  Purohit, Varshney, Kaza, Verma, Puri, Karia, Sampat, Doshi, Mishra, Reddy,
  Patro, Dixit, Shen, Baral, Choi, Smith, Hajishirzi, and
  Khashabi}]{Wang2022SuperNaturalInstructionsGV}
Yizhong Wang, Swaroop Mishra, Pegah Alipoormolabashi, Yeganeh Kordi, Amirreza
  Mirzaei, Anjana Arunkumar, Arjun Ashok, Arut~Selvan Dhanasekaran, Atharva
  Naik, David Stap, Eshaan Pathak, Giannis Karamanolakis, Haizhi~Gary Lai,
  Ishan Purohit, Ishani Mondal, Jacob Anderson, Kirby Kuznia, Krima Doshi,
  Maitreya Patel, Kuntal~Kumar Pal, M.~Moradshahi, Mihir Parmar, Mirali
  Purohit, Neeraj Varshney, Phani~Rohitha Kaza, Pulkit Verma, Ravsehaj~Singh
  Puri, Rushang Karia, Shailaja~Keyur Sampat, Savan Doshi, Siddharth~Deepak
  Mishra, Sujan Reddy, Sumanta Patro, Tanay Dixit, Xudong Shen, Chitta Baral,
  Yejin Choi, Noah~A. Smith, Hanna Hajishirzi, and Daniel Khashabi. 2022.
\newblock Super-naturalinstructions: Generalization via declarative
  instructions on 1600+ nlp tasks.
\newblock In \emph{Conference on Empirical Methods in Natural Language
  Processing}.

\bibitem[{Yang et~al.(2019)Yang, Feng, Qiao, Kan, and Li}]{Yang2019ExploringPL}
Sen Yang, Dawei Feng, Linbo Qiao, Zhigang Kan, and Dongsheng Li. 2019.
\newblock Exploring pre-trained language models for event extraction and
  generation.
\newblock In \emph{Annual Meeting of the Association for Computational
  Linguistics}.

\end{thebibliography}
\bibliographystyle{acl_natbib}

\end{document}